\documentclass{article}
\usepackage{spconf,amsmath,graphicx}
\usepackage{times}
\usepackage{epsfig}
\usepackage{graphicx}
\usepackage{amsmath}
\usepackage{amssymb}
\usepackage{booktabs}
\usepackage{bm}
\usepackage{multirow}
\usepackage{algorithm}  
\usepackage{algorithmic}  
\usepackage{bm}
\usepackage{setspace}
\usepackage{graphicx}
\usepackage{comment}
\usepackage{multicol}
\usepackage{wrapfig}
\usepackage{enumerate}
\usepackage{color}
\usepackage[colorlinks,urlcolor=violet]{hyperref}
\usepackage{array}
\usepackage{pifont}
\hypersetup{urlcolor=violet,
	allcolors=black,
	pdfstartview=Fit,
	breaklinks=true}

\def\mb{\mathbf}
\def\mbb{\mathbb}
\def\mc{\mathcal}

\def\mb{\mathbf}
\def\mbb{\mathbb}
\def\mc{\mathcal}

\def\ie{\textit{i.e.}}

\title{Centroid Distance Distillation for Effective Rehearsal\\in Continual Learning}
\name{Daofeng Liu$^{1*}$, Fan Lyu$^{2*}$\thanks{$*$ Co-first author.}, Linyan Li$^3$, Zhenping Xia$^1$, Fuyuan Hu$^{1**}$\thanks{$**$ Corresponding author: fuyuanhu@mail.usts.edu.cn.}\thanks{Supported by the Natural Science Foundation of China (No. 61876121).}}
\address{$^1$Suzhou University of Science and Technology,\\
	$^2$Tianjin University, $^3$Suzhou Institute of Trade \& Commerce}
\begin{document}
	%
\maketitle
\begin{abstract}
	Rehearsal, retraining on a stored small data subset of old tasks, has been proven effective in solving catastrophic forgetting in continual learning.
	However, due to the sampled data may have a large bias towards the original dataset, retraining them is susceptible to driving continual domain drift of old tasks in feature space, resulting in forgetting.
	In this paper, we focus on tackling the continual domain drift problem with centroid distance distillation. 
	First, we propose a centroid caching mechanism for sampling data points based on constructed centroids to reduce the sample bias in rehearsal.
	Then, we present a centroid distance distillation that only stores the centroid distance to reduce the continual domain drift.
	The experiments on four continual learning datasets show the superiority of the proposed method, and the continual domain drift can be reduced.
	Our code is available at
	\url{https://github.com/Daofeng-liu/CDD-R}.
\end{abstract}
\begin{keywords}
continual learning, distillation, centroids
\end{keywords}
\vspace{-5px}
\section{Introduction}


\vspace{-5px}
Continual Learning (CL) is used to enable a machine learning system to learn from a sequence of tasks like humans~\cite{review2019continual}, 
which has been applied to many applications, such as the recommender systems~\cite{recommender}, medical research~\cite{diagnosis,diagnostic} and clinical management decisions~\cite{clinical}.
However, CL suffers from a well-known obstacle in neural networks called catastrophic forgetting~\cite{mccloskey1989catastrophic}, which is the inability to effectively retain old knowledge after learning a new task. 
The objective of CL is to improve the adaptative ability to new knowledge over time without forgetting past knowledge.
To address catastrophic forgetting, the rehearsal method~\cite{8,LYU,Shim2021,6}, storing a small portion of data for replay, is proven simple and effective against other methods including regularization methods~\cite{5,hinton2015distilling,du2022agcn} and parameter-isolation methods~\cite{16,15,xu2018reinforced}.

\begin{figure}[t]
	\centering
	\includegraphics[width=\linewidth]{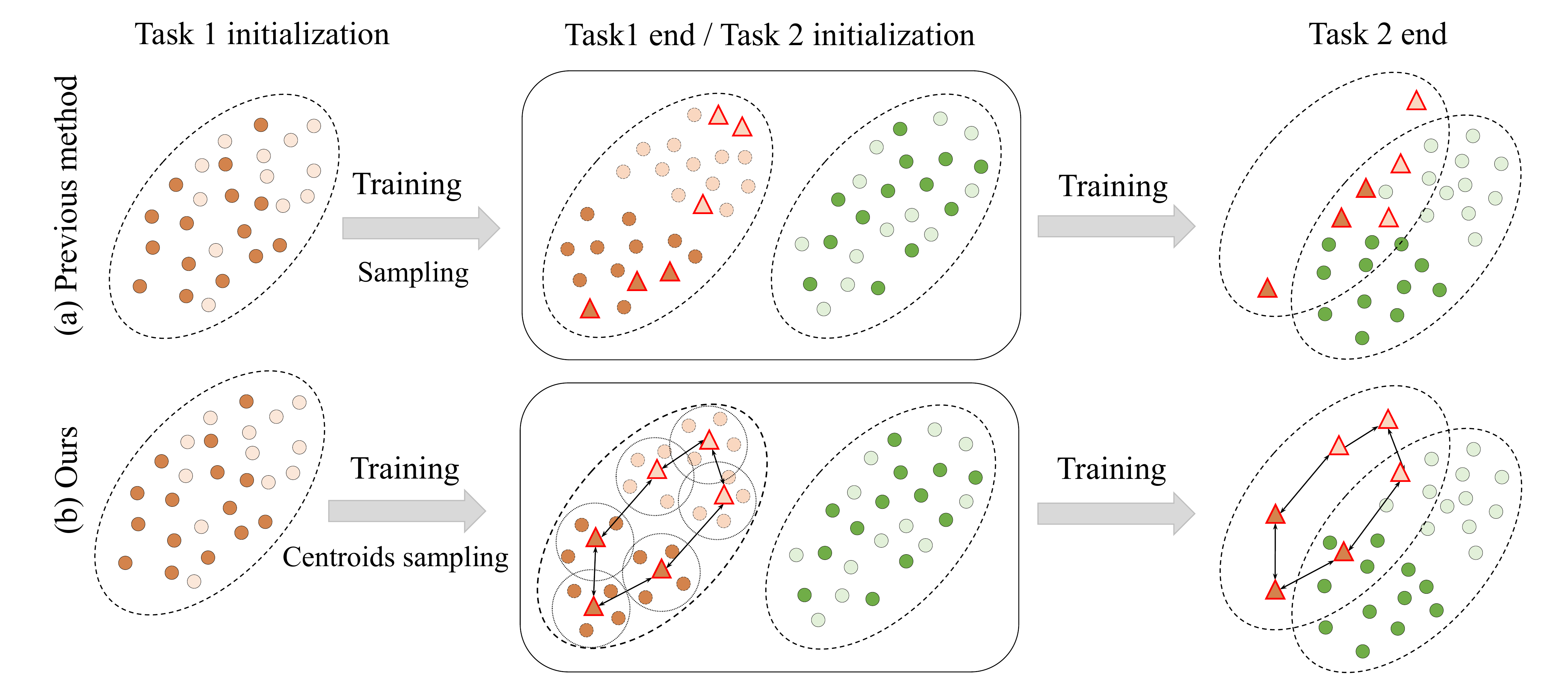}
	\vspace{-25px}
	\caption{
		(a) Previous rehearsal may select biased samples, which results in continual domain drift. 
		(b) Our rehearsal selects representative data based on centroids and suppresses continual domain drift through centroid distance distillation.
	}
	\label{fig:domainshift}
	\vspace{-15px}
\end{figure}

To reduce forgetting, as shown in Fig.~\ref{fig:domainshift}(a), the existing rehearsal may select biased samples for replaying and storing the corresponding features for distillation.
However, the biased samples have a poor capacity to represent the original domain.
Retraining on these samples leads to unpredictable drifts in feature space, namely, continual domain drifts (a.k.a. semantic drifts)~\cite{3}.
Moreover, distillation on the biased samples is memory-costly when the number of tasks grows, and the biased samples mislead the relative relationship between tasks, which results in over-fitting and indistinguishability of old tasks and slow learning of new tasks.



In this paper, we propose a simple yet effective Centroid Distance Distillation (CDD) for Rehearsal to mitigate the forgetting raised by continual domain drift.
As shown in Fig.~\ref{fig:domainshift}(b), CDD contains two main steps.
\textit{(1) Centroid-based sampling}:
we first select representative samples via a proposed centroid caching mechanism, which builds an auto-updated cache for each centroid for storing the most representative samples and the caches can also help the update of centroids.
\textit{(2) Centroid distance distillation}:
in contrast to distilling storage-costly features, we propose to only store the relative relationship, \ie, the pairwise centroid distance, which is distilled to guide the replay of old tasks for less forgetting with negligible storage.
We demonstrate that our method can better alleviate the catastrophic forgetting raised by continual domain drift with only stored centroid distance through the experiments on four popular CL datasets.



\vspace{-5px}
\section{Method}

\vspace{-5px}

\subsection{Preliminary: Continual domain drift in CL}


CL can be formulated as learning from a sequence of datasets $\left\{ {{\mc{D}_1},\cdots,{\mc{D}_T}} \right\}$ in order, where ${\mc{D}_t} = \left\{ {\left( {{x_{i}},{y_{i}}} \right)} \right\}_{i = 1}^{{N_t}}$ is the dataset for the $t$-th task.
We denote the model parameters as two main parts, the shared feature extraction layers $h: x\rightarrow {f}$ and the task-specific linear classification layer $g_t: {f}\rightarrow {p}$ for task $t$, where ${f}$ and ${p}$ are the deep feature and the predicted probability of the input image respectively.
In CL, the model suffers from catastrophic forgetting due to the unavailability of data from past tasks.
Rehearsal~\cite{8,tiwari2022gcr}, as an effective method to reduce forgetting, stores a small number of samples to memory $\mc{M}$, and the stored data will be retrained together with the current training to maintain the past task knowledge.

\begin{figure}[t]
	\centering
	\includegraphics[width=\linewidth]{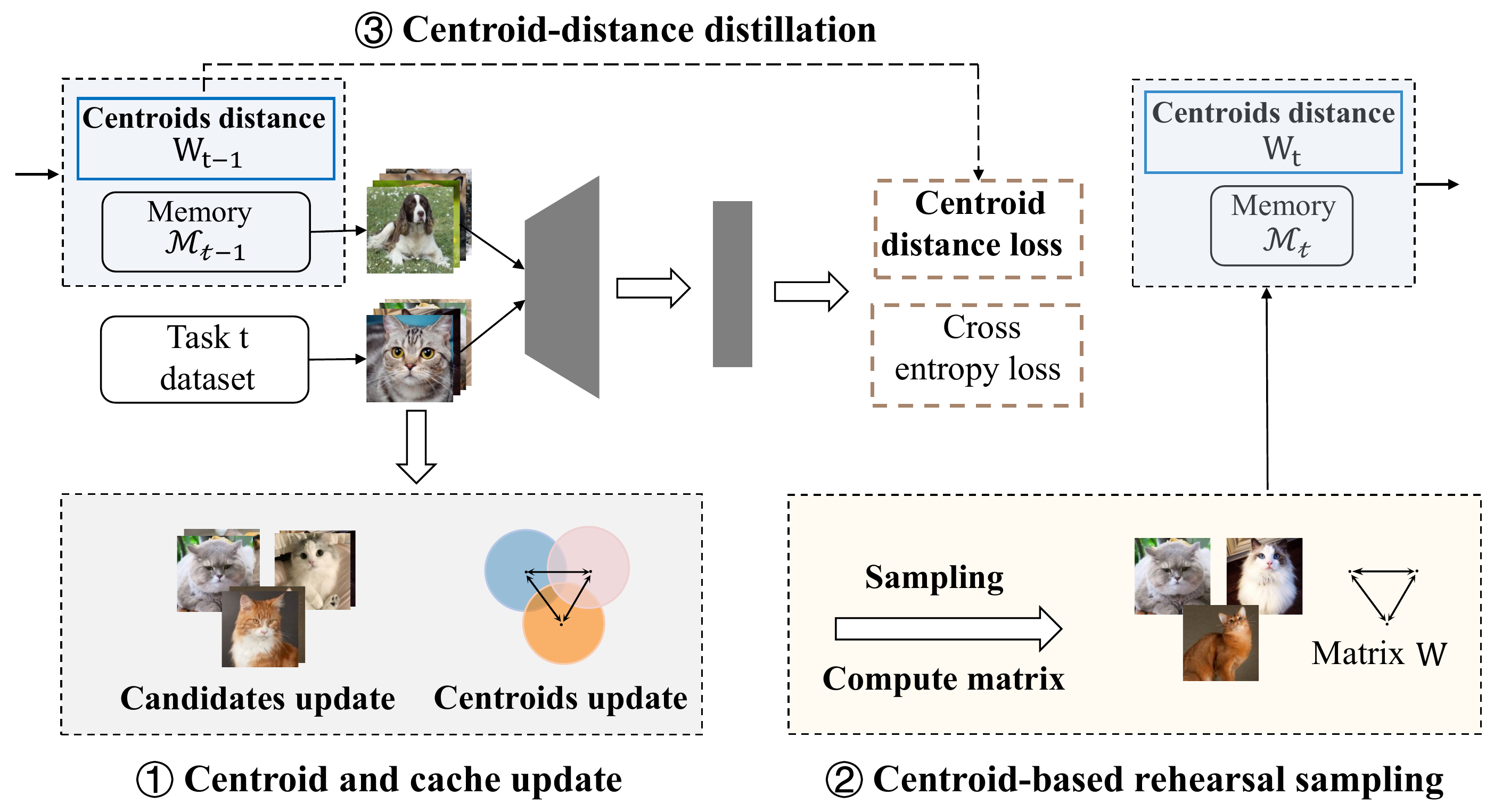}
	\vspace{-25px}
	\caption{
		{The framework of Centroid-Based Rehearsal.
		\large{\ding{172}} \normalsize The model computes a set of centroids and candidate samples at the cache layer. 
		\large{\ding{173}} \normalsize After task training, memory is selected from the cache by centroid-based rehearsal sampling.
		\large{\ding{174}} \normalsize  We compute the centroid relationship of memory data and distill the stored relationship to reduce continual domain drift. }
		}
	\label{fig:framework}
	\vspace{-15px}
\end{figure}

\vspace{-5px}
\subsection{Centroid-based Sampling for Rehearsal}

Although the continual learning method based on rehearsal improves the ability to remember past knowledge, the memory size is far small compared to the original data set $\mc{D}$, and the continual domain drift phenomenon will inevitably occur.
Moreover, the existing may sample data with large biases such as outliers, which even worse the continual domain drift.



To sample representative samples, following Agg-Var~\cite{14,ayub2020cognitively}, we propose a centroid-based sampling method for rehearsal.
A centroid $\mb{c}$ is the cluster center of the training domain~\cite{2020fewshot,centroid1,ayub2020cognitively}.
Given a training data point $(x,y)$, if the closest centroid $\mb{c}$ has a distance less than a given threshold $\epsilon$, this centroid is updated via:
\vspace{-5px}
\begin{equation}
	\vspace{-5px}
	{\mb{c}} = \frac{{{n} \times {\mb{c}} + h(x)}}{{{n} + 1}},
	\label{eq:agg}
\end{equation}
where ${n}$ is the number of the data points already represented by the centroid.
Elsewise, a new centroid will be constructed directly using $h(x)$.
We consider that the data pass only once and then discard in a stream following~\cite{Shim2021,tiwari2022gcr}.
It is difficult to determine which sample should be selected if they have been discarded before the optimal centroid.

\begin{algorithm}[t]
	\caption{{Update centroid and cache.}}
	\label{alg:centroid}

		\begin{algorithmic}[1]
			\renewcommand{\algorithmicrequire}{ \textbf{Input:}}
			\REQUIRE {Centroids} $\{\mb{c}_1,\cdots,\mb{c}_N\}$, {Centroid update numbers} $\{K_1,\cdots,K_N\}$, {model} $h$, {Cache} ${\mc{A}}$, {Distance threshold} $\epsilon$, {Cache size} $\gamma$, {data} $(x,y)$
			\renewcommand{\algorithmicrequire}{ \textbf{Output:}}
			\REQUIRE Updated $\{\mb{c}_1,\cdots,\mb{c}_{N'}\}$ and ${\mc{A}}$
			\STATE $i^* = \mathop{\arg\min}_{i\in[1,N]} \left\| h\left(x\right) - \mb{c}_{i} \right\|$ $\hfill\triangleleft$~\emph{Nearest centroid}	
			\STATE $d = \left\| h\left(x\right) - \mb{c}_{i^*} \right\|$
			\IF{$d>\epsilon$}
			\STATE $N'=N+1$, $\mb{c}_{N'} = h\left(x\right)$ $\hfill \triangleleft$~\emph{Create new centroid}
			\ELSE
			\STATE $N'=N$	
			\STATE $\mb{c}_{i^*}= 		
			\frac{{\sum\nolimits_{j = 1}^K h({x_j} \in{\mc{A}_{\mb{c}_{i^*}}})  + h(x)}}{{{K}_{i^* } + 1}}$ $\hfill \triangleleft$~\emph{Update centroid}	
			\ENDIF
			\IF{$K_{i^* } < \gamma$}
			\STATE ${\mc{A}_{\mb{c}_{i^*} }}\leftarrow{\mc{A}_{\mb{c}_{i^*} }}\cup{(x,y)}$ $\hfill\triangleleft$~\emph{Update cache}
			\ELSE
			\STATE $j = \mathop{\arg\max}_{j} \left\| {\mc{A}_{\mb{c}_{i^*}}} - \mb{c}_{i^*}
			\right\|$
			\STATE Remove ${\mc{A}_{\mb{c},j}}$ from ${\mc{A}_{\mb{c}}}$
			\ENDIF
		\end{algorithmic}
	\label{alg:centroid}

\end{algorithm}

Thus, together with the centroid updating, we build centroid-aware caches as the medium to select the closest sample candidates for each centroid, and the cache will be discarded after the task end.
For a centroid $\mb{c}$, its cache is represented as
${\mc{A}_{\mb{c}}} = \{ ({x_{j}},{y_{j}})\} _{j = 1}^K$.
The candidate data are the $K$ nearest data in the data stream to each centroid and are replaced with nearer samples in the current batch.
If the number of candidates for a centroid is less than a pre-defined $\gamma$, $(x, y)$ is added to the cache directly.
If the candidate number is equal to $\gamma$, the updated centroid compares the distance with $\gamma+1$ candidates and removes the farthest one.

After the training of the current task, we select the sample data from the cache into memory and delete the whole cache.
To prevent bias in the centroids and the updated model, in Eq.~\eqref{eq:agg}, we also consider using cache to update centroids:
\vspace{-5px}
\begin{equation}
\mb{c} = 		
\frac{{\sum_{x'\in\mc{A}_{\mb{c}}} h({x'})  + h(x)}}{{|\mc{A}_{\mb{c}}| + 1}},
\label{eq:centroid}
\vspace{-5px}
\end{equation}
where the centroid $\mb{c}$ is updated by the average of the features of the candidate data corresponding to this centroid (closest and distance less than $\epsilon$).
Easy to know, in Eq.~\eqref{eq:centroid}, important centroids are updated more frequently than some outliers. 
Based on the cache, the memory buffer is obtained by
\vspace{-5px}
\begin{equation}
	{\mc{M}^t} = \bigcup\nolimits^{|\mc{M}^t|}_i \left\{(x_i, y_i)\sim P^t({(x,y)})\right\},
\label{sampling}
\vspace{-5px}
\end{equation}
where $P^t_{i}= \frac{{{m_{i}}}}{{\sum\nolimits_{k = 1}^n {{m_{k}}} }}\in[0,1]$ is the sampling probability of the $i$-th centroid and $m_i$ is the total update frequency of centroid $i$.
It is worth noting why we do not only sample from caches with larger confidence because we need to also keep the diversity to some extent.
Otherwise, the stored samples will have larger biases than random selection

\begin{table*}[t]
	\centering
	\caption{
		Comparisons on four datasets, where the mean and std are over 5 seeds.
	}
	\label{table:headings}
	\setlength\tabcolsep{20pt}
	\resizebox{.9\linewidth}{!}{
		\begin{tabular}{l|ccc|ccc}
			\bottomrule
			\multirow{2}{*}{Methods} & \multicolumn{3}{c|}{\textbf{Permuted MNIST}} & \multicolumn{3}{c}{\textbf{Split CIFAR}}\\
			\cline{2-7}
			{} & ${A_T}(\% )$ & ${F_T}$ & $LTR$ & ${A_T}(\% )$ & ${F_T}$ & $LTR$\\
			\hline
			\noalign{\smallskip}
			A-GEM ~\cite{5}  & $89.32\pm0.46$ & $0.07\pm0.004$& $0.367\pm0.013$&  $61.28\pm1.88$ & $0.09\pm0.018$& $0.643\pm0.124$\\
			ER ~\cite{6}  & $90.47\pm0.14$ & $0.03\pm0.001$& $0.367\pm0.013$&  $63.97\pm1.30$ & $0.06\pm0.006$& $0.451\pm0.333$\\
			MEGA ~\cite{10}  & $91.21\pm0.10$ & $0.05\pm0.001$& $0.524\pm0.017$&  $66.12\pm1.94$ & $0.06\pm0.015$& $0.356\pm0.114$\\
			DER ~\cite{4}  &$92.03\pm0.19$& $0.04\pm0.001$&$ 0.402\pm0.012$&  $68.49\pm1.45$ & $0.06\pm0.009$& $0.371\pm0.087$\\
			ASER ~\cite{Shim2021} & - & - & - &$65.53\pm1.89 $&$ 0.07\pm0.007$&$ 0.544\pm0.133$\\
			SCR ~\cite{2} &$ 91.74\pm0.63$ & $ 0.05\pm0.004$&$ 0.492\pm0.041$&$ 67.99\pm1.89 $&$ 0.05\pm0.004$&$ 0.258\pm0.024$\\
			MDMTR ~\cite{3} &$91.97\pm0.23$&$ 0.05\pm0.002$ & $0.521\pm0.022$&$ 66.38\pm1.63  $&$ 0.05\pm0.006$&$ 0.377\pm0.076$
			\\
			MDMTR+FD ~\cite{3} &$93.97\pm0.15$&$ 0.03\pm0.002$ & $0.283\pm0.019$&$ 69.20\pm1.60  $&$ 0.04\pm0.010$&$ 0.283\pm0.099$
			\\\hline
			Ours &$92.22 \pm 0.22$&$0.04 \pm0.002$&$0.484\pm0.019$&$ 69.65\pm1.55 $&$ 0.04\pm0.013$&$ 0.192\pm0.094$\\
			Ours+FD &$\mb{94.12} \pm 0.11$&$\mb{0.01} \pm0.007$&$\mb{0.013}\pm0.009$&$ \mb{70.69}\pm2.33 $&$ \mb{0.03}\pm0.012$&$ \mb{0.119}\pm0.065$\\
			\midrule
			\multirow{2}{*}{Methods} & \multicolumn{3}{c|}{\textbf{Split CUB}} & \multicolumn{3}{c}{\textbf{Split AWA}}\\
			\cline{2-7}
			{} & ${A_T}(\% )$ & ${F_T}$ & $LTR$ & ${A_T}(\% )$ & ${F_T}$ & $LTR$\\
			\hline
			\noalign{\smallskip}
			A-GEM ~\cite{5}  & $61.82\pm3.72$ & $0.08\pm0.021$& $0.456\pm0.174$&  $44.95\pm2.97$ & $0.05\pm0.014$& $0.178\pm0.082$\\
			ER ~\cite{6}  & $73.63\pm0.52$ & $0.01\pm0.005$& $\mb{0.001}\pm0.001$&  $53.27\pm4.05$ & $0.02\pm0.030$& $0.014\pm0.015$\\
			MEGA ~\cite{10}  & $80.58\pm1.94$ & $0.01\pm0.017$& $0.002\pm0.002$&  $54.28\pm4.84$ & $0.05\pm0.040$ & $0.070\pm0.114$\\
			DER ~\cite{4}  & $76.56\pm2.48$ & $0.01\pm0.015$& $0.025\pm0.018$& $50.70\pm4.91$& $0.04\pm0.040$& $0.063\pm0.094$\\
			ASER ~\cite{Shim2021} & $ 75.58\pm3.72 $&$ 0.02\pm0.010$&$ 0.037\pm0.029$& $46.72\pm3.20$&$ 0.05\pm0.006$&$ 0.171\pm0.021$\\
			SCR ~\cite{2} & $ 81.43\pm1.97$&$ 0.01\pm0.007$&$ 0.007\pm0.009$&$ 54.35\pm2.68$ & $ 0.02\pm0.012$ & $ 0.022\pm0.010$\\
			MDMT ~\cite{3} & $ 83.06\pm4.39  $&$ 0.20\pm0.028 $&$ 0.015\pm0.023$ & $  58.20\pm2.51$ & $ 0.02\pm0.011$ & $ 0.035\pm0.025$
			\\
			MDMT+FD ~\cite{3} & $ 83.98\pm2.35  $&$ 0.01\pm0.015 $&$ 0.021\pm0.018$ & $  61.26\pm3.36$ & $ 0.02\pm0.027$ & $ \mb{0.002}\pm0.002$
			\\\hline
			Ours &$ 84.85\pm2.46 $&$ 0.01\pm0.012$&$ 0.008\pm0.010$&$59.26\pm4.72$&$ 0.03\pm0.037$&$ 0.034\pm0.065$\\
			Ours+FD &$ \mb{85.75}\pm1.99 $&$ \mb{0.01}\pm0.004$&$ 0.004\pm0.006$&$\mb{61.92}\pm2.94$&$ \mb{0.02}\pm0.027$&$ 0.013\pm0.002$\\
			\toprule
	\end{tabular}}
	\vspace{-20px}
\end{table*}

\vspace{-5px}
\subsection{Distillation on centroid distance }
Only selecting samples with diversity and representativeness for rehearsal based on the centroids is not enough to effectively solve the continual domain drift problem.
Because in the process of continual learning, domain drift will blur the decision boundary of the old task and cause forgetting of the old knowledge. 
A naive way is to anchor the memory feature as mentioned via distillation~\cite{3,4}.
However, constraining the memory feature move may undermine the new task learning and the stored features are a large storage burden.

In this paper, we propose to only store the relative relationships among centroids, i.e., the Cosine similarity.
In specific, we first calculate the similarity between any two centroids of the current task (intra- and inter-class),
and the pairwise distances can be represented by a matrix $\mb{W}^t$, where
\vspace{-5px}
\begin{equation}
	{\mb{W}_{ij}^t} = \frac{\mb{c}_i \cdot \mb{c}_j}{||\mb{c}_i||~ ||\mb{c}_j ||}.
\label{matrix}
\vspace{-5px}
\end{equation}
The storage cost of this matrix is much smaller than the raw samples and their features.


In the learning of new tasks, we recalculate the centroid distance matrix $\mb{W}^{t'}$ based on only the memory and distill it with the stored $\mb{W}^{t}$ with a Centroid Distillation (CD) loss 
\vspace{-5px}
\begin{equation}
{\mathcal{L}_\text{CD}} = \sum\nolimits_{t = 1}^{k - 1} \left\| {\mb{W}^t} - {\mb{W}^{t'}}\right\|^2.
\vspace{-5px}
\end{equation}
The centroid distance matrix is distilled in the following tasks to keep the decision boundary from blurring and drifting.
The CD loss takes advantage of the information between the centroids to further suppress the continual domain drift 
, where diversity and representativeness will be kept.
The relative relationships among centroids also have the classes of old tasks discriminative from each other in the current tasks.
\vspace{-5px}
\section{Experiments}

\begin{figure*}[t]
	\centering
	\includegraphics[width=.9\linewidth]{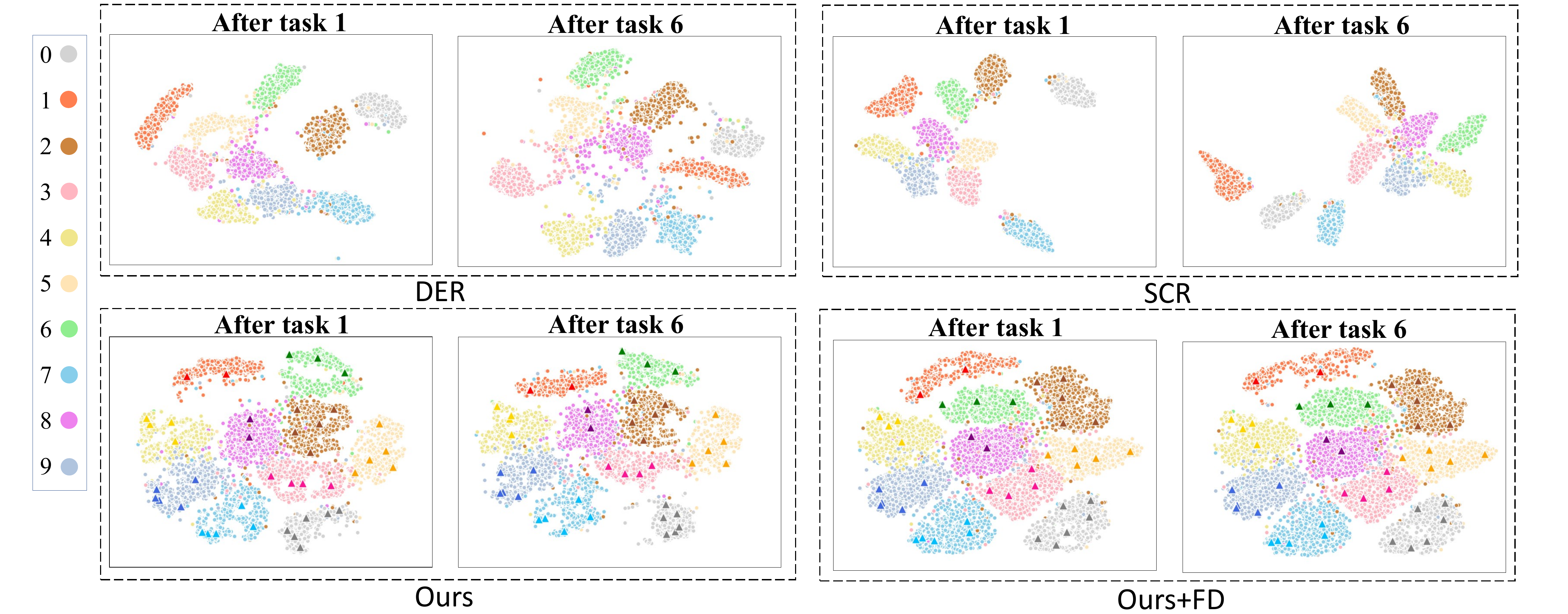}
	\vspace{-15px}
	\caption{
		T-SNE of task 1 on Permuted MNIST after the learning of task 1 and 6. Trangles are the centroids of each class.
	}
	\label{fig:tsne}
	\vspace{-15px}
\end{figure*}

\vspace{-5px}
\subsection{Dataset and experimental details}

\noindent
\textbf{Dataset.}
\emph{Permuted MNIST} is a variant of MNIST~\cite{lecun1998mnist} dataset where the input pixels for each task have different random permutation as different tasks. 
\emph{Split CIFAR} is a split version of the original CIFAR100 dataset~\cite{cifar100}, which splits 100 classes into 20 disjoint tasks, each containing 5 classes.
\emph{Split CUB} is a random splitting of the 200 classes of the CUB dataset~\cite{cub} into 20 disjoint tasks, each containing 10 classes.
\emph{Split AWA} is an incremental version of the AWA dataset~\cite{awa} that splits 50 animal categories into 20 tasks, each with 5 different categories, but the categories are repeatable across tasks.
We follow previous works~\cite{5,6,3} to conduct experiments on the above four datasets.
\\
\noindent
\textbf{Evaluation metric.}
\emph{Average Accuracy} (${A_T}$) is the average of the accuracy of all tasks after the model is trained on the last task.
\emph{Forgetting Measure} (${F_T}$) represents the the accuracy drop of old tasks after the model is trained on all tasks.
\emph{Long-Term Remembering} ($LTR$)~\cite{10} computes the accuracy drop of each task relative to when the task was first trained. 
\\
\noindent
\textbf{Implementation detail.}
Following~\cite{5,3}, we implement our methods with different backbones.
For Permuted MNIST, we use a fc network with two hidden layers of 256 ReLU units.
For Split CIFAR, we use a reduced resnet18~\cite{he2016deep}.
For Split CUB and Split AWA, we use a standard resnet18.
The model is optimized using stochastic gradient descent with a mini-batch size of 10.
For Permuted MNIST, Split CIFAR, Split CUB, Split AWA, $\epsilon$ and $\gamma$ are respectively set to (7,35) , (8,20) , (11,10) , (7,35) via grid searching.

\begin{table}[t]
	\centering
		\vspace{-7px}
		\caption{Ablation study on Split CIFAR.}		
		\resizebox{.8\linewidth}{!}{
			\begin{tabular}{c|c|ccc}
				\toprule
				 CD & FD & {$A_{\text{T}}(\%)$} & {$F_{\text{T}}$}  & {$LTR$}  \\
				\midrule
				 - & - & $66.38\pm1.63$ & $0.052\pm0.006$ &   $0.377\pm0.076$  \\
				 - & $\checkmark$ & $68.97\pm2.21$ & $0.040\pm0.009$ &   $0.254\pm0.031$  \\
				 $\checkmark$ & - & $69.65\pm1.55$ & $0.035\pm0.013$ & $0.192\pm0.094$  \\
				$\checkmark$ & $\checkmark$ & $\mb{70.69} \pm2.33$ & $\mb{0.027} \pm0.012$  & $\mb{0.119} \pm0.065$ \\
				\bottomrule
		\end{tabular}}
		\label{tab:abl}	
		\vspace{-20px}
\end{table}

\vspace{-5px}
\subsection{Experimental Results}

\noindent
\textbf{Main comparisons.}
As shown in Table~\ref{table:headings},  our method compares with other SOTAs\cite{9,5,6,10,4,3,Shim2021,2} in three metrics.
For ${A_T}$, our method shows a clear improvement compared to other methods on all four datasets. 
This indicates that our method can better suppress forgetting and reduce domain drift on all seen tasks.
For ${F_T}$ and $LTR$, our method also has an advantageous position, which shows that our method can gain long-term memory by reducing the domain drift.
Our method achieves the best results compared to SOTAs without FD (Feature Distillation) loss~\cite{3}. 
With FD loss, our method can be further improved.

\begin{table}[t]
	\centering
	\vspace{-5px}
	\caption{The effect of the centroid distance $\epsilon$.}		
	\resizebox{.75\linewidth}{!}{
		\begin{tabular}{c|ccc}
			\toprule
			$\epsilon$ & {$A_{\text{T}}(\%)$} & {$F_{\text{T}}$}  & {$LTR$}  \\
			\midrule
			6 &   $69.76\pm1.59$ & $0.034\pm0.007$  & $0.172\pm0.044$ \\
			7 &   $70.16\pm1.96$ & $0.029\pm0.012$  & $0.148\pm0.072$ \\
			8 & $\mb{70.69} \pm2.33$ & $\mb{0.027} \pm0.012$  & $\mb{0.119} \pm0.065$ \\
			9  & $69.81\pm1.76$ & $0.030\pm0.008$  & $0.143\pm0.048$ \\
			
			\bottomrule
		\end{tabular}}
	\label{tab:ab2}	
	\vspace{-15px}
\end{table}

\begin{table}[t]
	\centering
	 \vspace{-7px}
	\caption{Sampling comparisons on Split CIFAR. }
	\resizebox{.8\linewidth}{!}{
		\begin{tabular}{c|ccc}
			\toprule
			Methods (without distillation)& {$A_{\text{T}}(\%)$} & {$F_{\text{T}}$}  & {$LTR$}  \\
			\midrule
			 Ring buffer & $66.38\pm1.63$ & $0.052\pm0.006$ &   $0.377\pm0.076$  \\
			 MoF & $66.58\pm1.75$ & $0.053\pm0.010$ & $0.359\pm0.106$  \\
			 GSS & $62.06\pm3.58$ & $0.115\pm0.021$  & $0.912\pm0.183$ \\
			 \midrule
			 Ours & $\mb{69.18}\pm0.74$ & $\mb{0.039}\pm0.008$ & $\mb{0.236}\pm0.061$ \\
						
			\bottomrule
	\end{tabular}}
	\label{tab:ab3}	
	\vspace{-20px}
\end{table}

\begin{figure}[t]
	\centering
	\includegraphics[width=.79\linewidth]{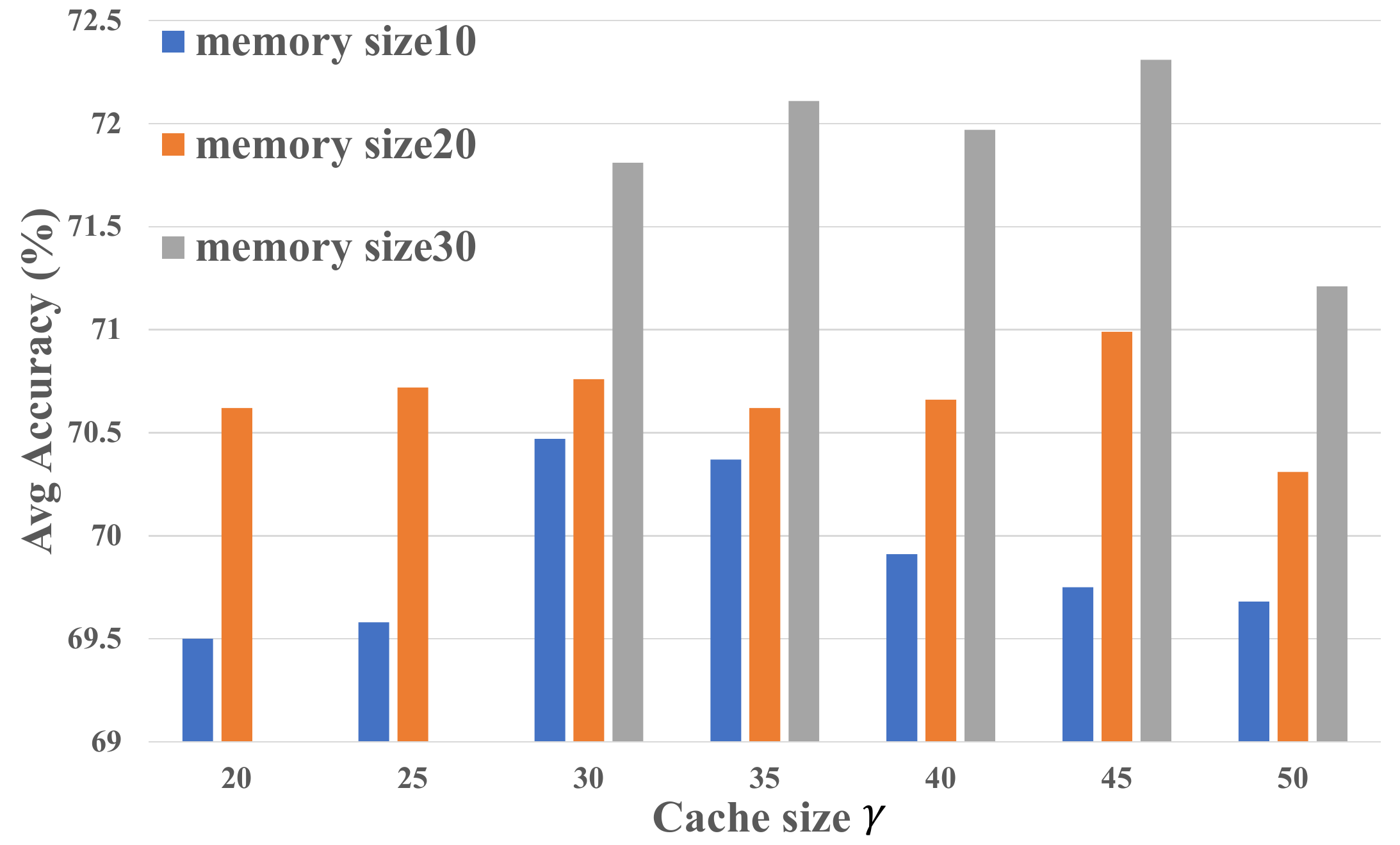}
	\vspace{-15px}
	\caption{
		The effect of the cache size $\gamma$ with different memory size settings.
		Memory size is the number of samples per class.
	}
	\label{fig:cache}
	\vspace{-15px}
\end{figure}

\noindent
\textbf{Ablation study.}
We then explore the impact of each configuration. 
We show the ablation experiments on the {Split CIFAR} dataset in Table~\ref{tab:abl}.
The first row of the table shows the performance of our experimental baseline, without adding any configuration.
When we add CD loss or FD loss, the experimental results can be clearly improved. 
The experimental performance is further improved when CD loss and FD loss are used together.
In Table~\ref{tab:ab2}, We also conduct an experimental study on the choice of the hyper-parameter {centroid distance $\epsilon$}.
The larger the centroid distance, the less the class trains out of the centroid, and the smaller the centroid distance, the more the centroid is obtained. 
Easy to observe, when a suitable centroid distance is chosen, the experimental effect can reach optimal performance.
In Fig.~\ref{fig:cache}, we set three memory sizes on the Split CIFAR dataset to
explore the effect of changing the cache size $\gamma$ on the experimental results. 
We can see that for different memory sizes, the optimal cache sizes are different.
Moreover, a larger memory size always means better performance with an appropriate cache size.

\noindent
\textbf{Sampling strategy comparisons.}
In Table~\ref{tab:ab3}, we compare with other sampling methods. 
Ring buffer is classical online random sampling methods. 
Mean-of-Feature (MoF)~\cite{8} samples the data closest to the mean by calculating the mean of the features for all data in each class, which means larger storage is needed.
Gradient-Based Sample Selection (GSS)~\cite{gradient} diversifies the gradients of the samples in the memory buffer.
In contrast, our centroid-based sampling achieves the best results without much storage.

\noindent
\textbf{Continual domain drift observation}
We explore the continual domain drift phenomenon in continual learning by the t-SNE visualization.
In Fig.~\ref{fig:tsne}, we show the features distribution of Task 1 on Permuted MNIST at the end of Task 1 phase and Task 6 phase.
The DER and SCR methods do not suppress the occurrence of continual domain drift and thus lead to the forgetting of past knowledge.
Our method can effectively mitigate the continual domain drift by centroid-based sampling and centroid distance distillation, and the feature distribution remains relatively stable after the end of Task 6.
When our method is combined with FD loss, the continual domain drift phenomenon is further reduced.

\vspace{-5px}
\section{Conclusion}
\vspace{-5px}
In this paper, we tackled catastrophic forgetting from the perspective of continual domain drift.
We first constructed centroids for each class in an online fashion.
Then, with guidance from centroids, we stored representative data points to reduce the dataset bias.
We also stored the relative centroid distance, which is used to distill for long-term remembering.
The experimental results on four continual learning datasets show the superiority of our method.

\newpage
\small
\bibliographystyle{IEEEbib}	
\bibliography{refs}

\end{document}